\pdfoutput=1

\documentclass[11pt, dvipsnames]{article}

\usepackage[]{EMNLP2022}

\usepackage{times}
\usepackage{latexsym}

\usepackage[T1]{fontenc}

\usepackage[utf8]{inputenc}
\usepackage{graphicx}
\usepackage{amsmath}
\usepackage[ruled, lined, longend, linesnumbered]{algorithm2e}
\usepackage{floatrow}
\usepackage{colortbl}
\usepackage{todonotes}
\usepackage{subcaption}
\usepackage{pdfpages}
\usepackage{hyperref}
\usepackage{array}
\usepackage{multirow}

\usepackage{microtype}

\usepackage{inconsolata}

\newfloatcommand{capbtabbox}{table}[][\FBwidth]
\newcommand{\len}{LEN$^p$}

\title{Extending Logic Explained Networks to Text Classification}

\author{
  Rishabh Jain\thanks{\phantom{ a}Corresponding author: \texttt{rj412@cam.ac.uk}} \\
  University of Cambridge\\
  Cambridge, UK\\
  \And
  Gabriele Ciravegna \\
  Université Côte d’Azur, Inria,\\ CNRS, I3S, Maasai, Nice, France\\  
  \And
  Pietro Barbiero \\
  University of Cambridge\\
  Cambridge, UK \\
  \AND
  Francesco Giannini \\
  University of Siena\\
  Siena, Italy\\
  \And
  Davide Buffelli \\
  University of Padova \\
  Padova, Italy \\
  \And
  Pietro Lio \\
  University of Cambridge\\
  Cambridge, UK \\
  }

\begin{document}
\maketitle

\begin{abstract}
Recently, Logic Explained Networks (LENs) have been proposed as 
explainable-by-design neural models providing logic explanations for their predictions.
However, these models have only been applied to vision and tabular data, and they mostly favour the generation of global explanations, while local ones tend to be noisy and verbose.
For these reasons, 
we propose \len{}, improving local explanations by perturbing input words, and we test it on text classification. 
Our results show that (i) \len{} provides better local explanations than LIME in terms of sensitivity and faithfulness, and (ii) logic explanations are more useful and  user-friendly than feature scoring provided by LIME as attested by a human survey. 
\end{abstract}

\section{Introduction}

The development of Deep Neural Networks has enabled the creation of high accuracy text classifiers \citep{lecun2015deep} with state-of-the-art models leveraging different forms of architectures, like RNNs (GRU, LSTM) \citep{text-classification-dnns} or Transformer models \citep{vaswani2017attention}.  However, these architectures are considered as black-box models \citep{adadi2018peeking}, since their decision processes are not easy to explain and depend on a very large set of parameters.
In order to shed light on neural models' decision processes, eXplainable Artificial Intelligence (XAI) techniques attempt to understand text attribution to certain classes, for instance by using white-box models.
Interpretable-by-design models engender higher trust in human users with respect to explanation methods for black-boxes, at the cost, however, of lower prediction performance. 

Recently,  \citet{ciravegna2021logic} and \citet{ barbiero2022entropy} introduced the Logic Explained Network (LEN), an explainable-by-design neural network combining interpretability of white-box models with high performance of neural networks. 
However, the authors only compared LENs with white-box models and on tabular/computer vision tasks. 
For this reason, in this work we apply an improved version of the LEN (\len{}) to the text classification problem, and we compare it with LIME \citep{lime}, a 
very-well known explanation method.  
LEN and LIME provide different kind of explanations, respectively FOL formulae and feature-importance vectors, and we assess their user-friendliness by means of a user-study.  
As an evaluation benchmark, we considered Multi-Label Text Classification for the 
tag classification task on the \href{https://www.kaggle.com/stackoverflow/stacksample}{``StackSample: 10\% of Stack Overflow Q\&A''} dataset \citep{stacksample}.

\textbf{Contribution}
The paper aims to: 
        (i) improve LEN explanation algorithm (\len{}{})\footnote{\len{} has been integrated in the original LEN package, and it is available at: \url{https://pypi.org/project/torch-explain/}};
        (ii) compare the faithfulness and the sensitivity of the explanations provided by LENs and LIME; 
        (iii) assess the user-friendliness of the two kinds of explanations.

\section{Background}


\paragraph{Explainable AI}
Explainable AI (XAI) algorithms describe the rationale behind the decision process of AI models in a way that can be understood by humans. 
Explainability is essential in increasing the trust in the AI model decisions, as well as in providing the social right to explanation to end users \citep{right-to-exp}, especially in safety-critical domains. Common methods include LIME 
\citep{lime}, SHAP 
\citep{shap}, LORE 
\citep{guidotti2018local}, 
Anchors \citep{ribeiro2018anchors} and many others.  

\begin{figure}[t]
        \centering
        \includegraphics[width=1\linewidth]{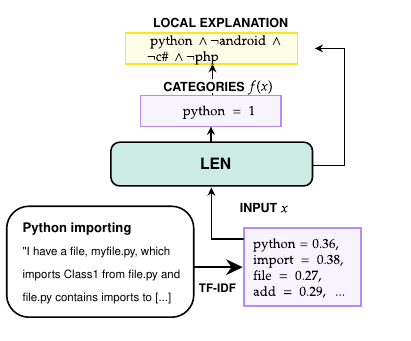}
    \caption{Example of LEN local explanations for a text predicted as `python' tag. }
    \label{img:len}
\end{figure}


\paragraph{LEN}
\label{sec:LEN-intro}
The Logic Explained Network \citep{ciravegna2021logic} is a novel XAI architectural framework forming special kind of neural networks that are explainable-by-design. In particular, LENs impose architectural sparsity to provide explanations for a given classification task. Explanations are in the form of First-Order Logic (FOL) formulas approximating the behaviour of the whole network. 
A LEN $f$ is a mapping from $[0,1]^d$-valued input concepts to $r\geq 1$ output explanations, that can be used either to directly classify data and provide relevant explanations or to explain an existing black-box classifier. 
At test time, a prediction $f_i(x) = 1$ is locally explained  
by the conjunction $\phi^{(i)}_{l}$ of the most relevant input features for the class $i\in\{1,\ldots,r\}$: 
\vskip -0.4cm
\begin{equation}
    \text{LEN Local Exp.: } \phi_{l}^{(i)}(x) = \bigwedge_{\mathbf{x_j} \in \mathcal{A}^{(i)}} \mathbf{x_j}(x),
\end{equation}
\vskip -0.2cm
\noindent
where $\mathbf{x_j}(x)$ is a logic predicate associated to the $j$-th input data and $\mathcal{A}^{(i)}$ is the set of relevant input features for the $i$-th task. Any $\mathbf{x_j}(x)$ can be either a positive $\mathbf{x_j}(x)$ or negative $\neg \mathbf{x_j}(x)$ literal, according to a given threshold, e.g. $\mathbf{x_j}(x) = [x_j > 0.5]$.
In this work, we consider the LEN proposed in \citet{barbiero2022entropy}, where the set of important features $\mathcal{A}^{(i)}$ for task $i$ is defined as $\mathcal{A}^{(i)} = \{\mathbf{x_j} \ | \ 1 \leq j \leq d,\ \alpha_j^{(i)} \geq 0.5\}$, where 
$\alpha_j^{(i)}$ is the importance score of the $j$-th feature, 
computed as the normalized softmax over the input weights $W$ connecting the $j$-th input to the first network layer
$||W^{(i)}_j||_1$. Architectural sparsity is obtained by minimizing the entropy of the $\alpha$ distribution. 

For global explanations, LENs consider the disjunction of the most important local explanations:
\vskip -0.4cm
\begin{equation}
    \text{LEN Global Exp.: } \phi_{g}^{(i)} = \bigvee_{\phi_{l}^{(i)} \in \mathcal{B}^{(i)}} \phi_{l}^{(i)},
\end{equation}
\vskip -0.2cm
\noindent
where $\mathcal{B}^{(i)}$ collects the $k$-most frequent local explanations of the training set and is computed as $\mathcal{B}^{(i)} = \{\phi_{l}^{(i)} \in  \arg\max_{\phi_{l}^{(i)} \in \Phi_{l}^{(i)}}^{k} \mu(\phi_{l}^{(i)}) \}$, where we indicated with $\mu(\cdot)$ the frequency counting operator and with $\Phi_{l}^{(i)}$ the overall set of local explanations related to the $i$-th class. In addition to this, \citet{ciravegna2021logic} employs a greedy strategy,
gradually aggregating frequent local explanations only if they improve the validation accuracy. 


\section{\len{}}


\subsection{Improving Local Explanation}
\label{sec:local-exp-imp}
The LEN algorithm for obtaining local explanations is not precise in determining the contribute of each feature. A close look at the extraction method shows that 
the $\alpha$ score only highlights the importance of a feature, 
without considering the type of contribution (either positive or negative) 
for the predicted class. 
As an example, consider an input text 
predicted as referring to \verb|C#|. The LEN may have learned that the presence of the word \verb|C#| 
leads to the tag prediction \verb|C#| and so it has assigned a high importance value $\alpha_{\small\verb|C#|}^{\small(\verb|C#|)}$. 
However, sometimes we may not have the word \verb|C#| in the text and still get the prediction to be \verb|C#|. 
The algorithm proposed in \citep{barbiero2022entropy} would extract a local explanation with the term $\neg \verb|C#|$, as shown in Figure \ref{fig:sample-local-exps}.
This is inaccurate because the absence of \verb|C#| does not lead to prediction of the tag \verb|C#|. 


To improve the local explanations of LENs, we take the most important terms $\mathcal{A}^{(i)}$ and 
we divide them into two subsets -- the \emph{good terms} and the \emph{bad terms}. The \emph{good terms} are the ones that \emph{actually} lead to the prediction. The \emph{bad terms} are the ones \emph{despite} which we get the given prediction.
For each term, we decide whether it is good or bad by comparing the predicted probability of the tag with the current input and with a perturbed
one (flipping term presence). 
If the prediction increases with the perturbation, the term is labelled as a bad term, otherwise it is considered a good term. Notice that the logic sign still comes from the input feature presence/absence. For ease, we only consider the conjunction of the good terms as the final explanation. Figure \ref{fig:sample-local-exps} shows the ability of \len{}
local explanation algorithm to correctly identify that the prediction is despite the absence of \verb|C#|.
Algorithm \ref{alg:local-exp-imp} in Appendix \ref{app:alg}, shows the pseudocode for the \len{} local explanations.

We note that a similar algorithm is used 
in Anchors \citep{ribeiro2018anchors}. However, our approach can be more effective as we perturb and assess the importance of both the given input words and of the (important) absent ones. Indeed, Anchors formulae only report positive literals by inspecting the global behaviour of the model, while we also provide logic explanations with negative terms. 


\begin{figure}[tbh]
\centering
\caption{Sample local explanations using the original LEN strategy and the improved \len{} strategy. The model has learned that C\#, .NET in input leads to tag C\# and
have high importance.}
\label{fig:sample-local-exps}
\small
\begin{tabular}{|p{0.2\linewidth}|p{0.6\linewidth}|}
      \hline
      Question & Which .NET collection should I use for adding multiple objects at once and getting notified? \\
      \hline
      Predicted Tags & C\# \\
      \hline
      \multirow{2}{0.2\linewidth}{LEN explanation} & $\neg$ C\# $\wedge$ .NET \\
      & \\
      \hline
      \multirow{2}{0.2\linewidth}{\len{} explanation} & .NET \\
      & ($\neg$ C\# is a bad term, discarded)\\
      \hline
\end{tabular}
\end{figure}

\subsection{Global Explanation}
\label{sec:global-exp-imp}
The greedy aggregation technique in \citet{ciravegna2021logic} 
may not find an optimal solution. 
The time complexity of the original aggregation method is $O(k \times n)$, since they evaluate the validation accuracy of the global formula (for $n$ samples) while aggregating the $k$ local explanations.
However, when we aggregate a small number of local explanations, i.e. $k$ is small, we can afford a more effective but slower solution.
Straightforwardly, we compute the disjunctions of all the possible combinations of local explanations (power set), incurring in a $O(2^k \times n)$ time complexity, but finding an optimal solution, i.e. the one reaching the higher validation accuracy.
Note that to keep the explanations short and easy to interpret, normally $k$ is very small, between $3$ and $10$. 
In Appendix \ref{app:alg}, Algorithm \ref{alg:global-exp-imp} shows the improved \len{} aggregation method. 


\section{Experiments}
In the experimental section, we show that 
(i) \len{} improves LEN explanations and provides better explanations than LIME in terms of faithfulness, sensitivity and capability to detect biased-model (Section \ref{sec:exp_evaluation}) and (ii) a human study confirms this result, in particular when considering the global explanation (Section \ref{sec:eval-survey}). Furthermore, in Appendix \ref{sec:model-eval}, we confirm that LENs achieve competitive performance when employed as explainable-by-design classifier w.r.t. black-box models. Appendix~\ref{sec:details} contains experimental details.

\subsection{Explanation Comparison}
\label{sec:exp_evaluation}
To assess the quality of the explanations of the proposed method (\len{}), we compared it with the original LEN algorithm (LEN), a version of LIME with discretized input (LIME (D)), and a version of LIME with non-discretized input (LIME (ND))\footnote{Both LIME (D) and LIME (ND) are provided in the LIME package \url{https://pypi.org/project/lime/}.}. 

We compare the different strategies by explaining a common black-box Random Forest model. Due to the high computational complexity required to explain each of the $\sim$ 15K tags (reduced from the initial 37K tags, after retaining only important questions), 
we compare the local explanations overs three tags only, namely ``C\#'', ``Java'' and ``Python''.
The hyperparameters of each method were chosen to get the best results while keeping the computational time to be at most 15 minutes.

\paragraph{\len{} provides faithful explanations}
The faithfulness of an explanation to a model refers to how accurate the explanation is in describing the model decision process. 
To evaluate the faithfulness, we use the Area Under the Most Relevant First perturbation Curve (AUC-MoRF). The lesser the value of AUC-MoRF, the more faithful is the explanation to the model. 
We calculate the AUC-MoRF for each strategy, 
considering the local explanation over 100 samples.
\begin{table}[t]
\centering
\caption{Average AUC-MoRF with 95\% confidence interval. The lower, the better. }
\label{table:auc-morf}
\small
\begin{tabular}{|c|c|}
      \hline
      \bfseries Explanation & \bfseries AUC-MoRF \\
      \bfseries Strategy & \\
      \hline
      \hline
      LEN & $0.4985 \pm 0.0283$ \\
      \hline
      \rowcolor{lightgray}
      \len{} & $0.0489 \pm 0.0117$ \\
      \hline
      LIME (D) & $0.4413 \pm 0.0171$ \\
      \hline
      LIME (ND) & $0.3919 \pm 0.0159$ \\
      \hline
\end{tabular}
\end{table}
\begin{table}[t]
\centering
\caption{Average Max-Sensitivity with 95\% confidence interval.  The lower, the better.}
\label{table:max-sens}
\small
\begin{tabular}{|c|c|}
      \hline
      \bfseries Explanation & \bfseries Max-Sensitivity \\
      \bfseries Strategy & \\
      \hline
      \hline
      \rowcolor{lightgray}
      LEN & $0.0000 \pm 0.0000$ \\
      \hline
      \rowcolor{lightgray}
      \len{} & $0.0000 \pm 0.0000$ \\
      \hline
      LIME (D) & $1.4031 \pm 0.1482$ \\
      \hline
      LIME (ND) & $1.3978 \pm 0.0467$ \\
      \hline
\end{tabular}
\end{table}
\begin{table}[t]
\centering
\caption{Capability to detect biased-model. We report the percentage of times the compared explanation algorithm detect the use of noisy features in biased model in two experimental settings.}
\label{table:sim-exp}
\small
\begin{tabular}{|c|c|c|}
      \hline
      \bfseries Explanation & \bfseries S1 & \bfseries S2 \\
      \bfseries Strategy & & \\
      \hline
      \rowcolor{lightgray}
      \len{} & $80\%$ & $95\%$ \\
      \hline
      LEN & $10\%$ & $55\%$ \\
      \hline
      SP-LIME (D) & $05\%$ & $25\%$ \\
      \hline
      SP-LIME (ND) & $00\%$ & $00\%$ \\
      \hline
\end{tabular}
\end{table}

Table \ref{table:auc-morf} reports the average AUC-MoRF for the different explanation strategies. The \len{} provides more faithful explanations than all the competitors by a considerable margin. On the contrary, the original LEN explanations are slightly less faithful than LIME (D) and LIME (ND).

\paragraph{\len{} explanations are robust to perturbations}
The sensitivity of an explanation refers to the tendency of the explanation to change with minor changes to the input. In general, a robust explanation is not affected by small random perturbations, since we expect similar inputs to have similar explanations. Therefore, low sensitivity values are desirable and we measure the Max-Sensitivity. For more details about the metric, please refer to Appendix \ref{sec:max-sens}.
Table \ref{table:max-sens} report the average Max-Sensitivity evaluated over 100 randomly selected inputs and performing 10 random perturbations $x^\star$ per input $x$, with maximum radius $\epsilon=||x - x^\star||_\infty=0.02$.
We see that explanations from both LEN and \len{} have $0.0$ Max-Sensitivity, i.e., they remain unchanged by all minor perturbations to the input, greatly outperforming the explanations from LIME. This is expected because LEN trains the model once over all the training data and tries to act as a surrogate model; there is no retraining for a new local explanation. 
On the other hand, LIME trains a new linear model for each local explanation and only try to mimic the explained model on the predictions near the given input. 
Clearly, by employing larger perturbation of the input, LEN explanations would also change.


\paragraph{\len{} is capable to detect biased-model} \label{sec:sim-exps}
The presence of noisy features in the training data may drive a model to unforeseeable prediction on clean data at test time. For this reason, it is very important to detect them before releasing the model. A way to detect biases is to compute the global explanation of a model and check whether the explanation is consistent with the domain knowledge. To this aim, it is very important to employ a powerful explanation algorithm that may be capable to detect the bias.
To evaluate this capability, we trained a model with the explicit goal of making it biased. 
In the training data, we added noisy features with a high correlation with certain tags, so that the model learns to associate the noisy features with the tag. 
At test time, these features are added randomly, i.e. they act like noise. We run these experiments in two settings, S1 and S2, varying the amount of bias towards the noisy features. 
This is done by increasing the bias of the noisy features in training data from 30\% of training data in S1 to 35\% in S2, and by ensuring a higher difference in test and validation scores in S2. 

Table \ref{table:sim-exp} reports the percentage of times we are able to detect the use of noisy features using the global explanations from the different strategies. \len{} shows more utility than all
the competitors by a large margin. The results have been averaged over 20 executions in this setting.

\subsection{Human Survey}
\label{sec:eval-survey}
We carried out a human survey to compare the ease of understanding and the utility of the explanations obtained by LIME and LENs.
The human survey was approved by the ethics committee and the questions do not record personal information.
The survey was shared with students and researchers over different universities 
and filled by $26$ respondents, $13$ with experience in Machine Learning, $10$ in Computer Science and $3$ in neither.
The survey is attached in Appendix \ref{sec:appendix-survey}. 
Figure \ref{plot:survey} 
report the ease of using the explanations for the different task. 


\begin{figure}[t]
      \includegraphics[width=0.9\linewidth]{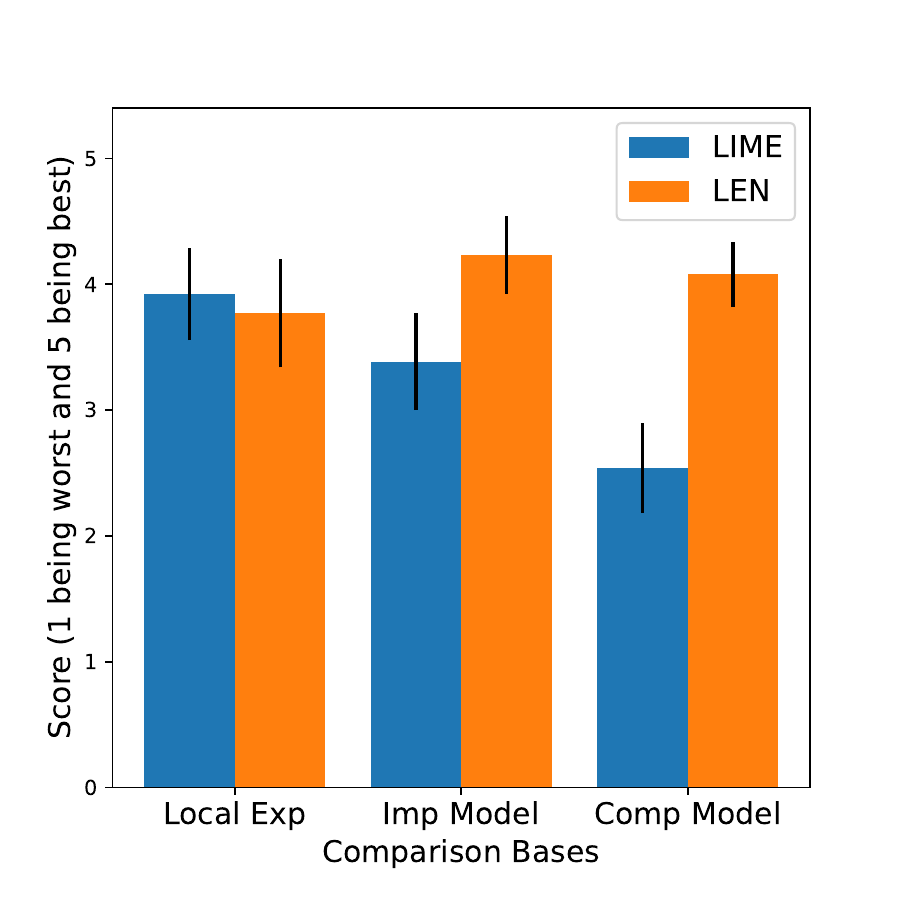}
      \caption{Survey rating results with $95\%$ confidence intervals}
      \label{plot:survey}
\end{figure}

\begin{table}[t]
    \centering
    \caption{Survey tasks results}
    \label{table:survey-task-eval}
    \small
    \begin{tabular}{|p{0.5\linewidth}|c|c|}
        \hline
        \bfseries Quantity Measured & \bfseries LIME & \bfseries LEN \\
        \hline
        \ & & \\
        Respondents able to identify the feature to ignore to improve the classifier & 
        \hfil$61.5\%$\hfil
        &
        \hfil$84.6\%$\hfil
        \\
        \ & & \\
        \hline
        \ & & \\
        Respondents able to identify the more general classifier 
        &
        \hfil$50.0\%$\hfil
        &
        \hfil$73.1\%$\hfil
        \\
        \ & & \\
      \hline
    \end{tabular}
\end{table}

\paragraph{\len{} explanations are easily interpretable}
\label{sec:survey-local}
First, the survey presents
the respondents a sample input 
with the related prediction and the explanations from LIME and \len{}. It then asks the respondents to rate the ease of understanding these local explanations. The first column of Figure \ref{plot:survey} suggests that the local explanations from LIME and \len{} are almost equally easily understandable 
since the $95\%$ confidence intervals have a high overlap (with LIME having a slightly higher mean).

\paragraph{\len{} enable users to improve a classifier}
\label{sec:survey-imp-clf}
This section aims to establish the usefulness of the explainers in getting a more general classifier through feature engineering. 
To this aim, we trained a radial basis function (RBF) SVM to mistakenly learn to associate the feature `\verb|add|' with the \verb|C#| tag. This was done by perturbing the training data and ensuring a wide difference in the validation and testing scores. We asked respondents to identify the input feature which does not allow the model to generalize well, by inspecting the global explanation of SP-LIME and \len{}. 
They were also asked to rate the ease of using the explanations for this purpose.
As shown in Table \ref{table:survey-task-eval} first row, only $\textbf{61.5\%}$ respondents were able to identify the term `\verb|add|' as the feature to ignore while using LIME explanations, as opposed to \len{} $\textbf{84.6\%}$.
In addition, in Figure \ref{plot:survey} second column, respondents found \len{} easier to use for improving the classifier. 

\paragraph{\len{} allows identifying the best classifiers}
\label{sec:survey-compare-clfs}
Finally, we evaluated whether users can 
choose a classifier that generalizes better than the other, 
by only checking again the global explanations 
of the two classifiers. 
To this aim, we trained two RBF SVMs classifiers on different training data (the second one with some noise added).
As reported in Table \ref{table:survey-task-eval}, second row, $\textbf{73.1\%}$ respondents were able to identify the more general classifier using \len{} as opposed to LIME $\textbf{50\%}$.
Moreover, the third column of Figure \ref{plot:survey} shows that the global explanations from \len{} make the comparison much easier than those from SP-LIME. 


\section{Conclusion}
This paper proposes \len{}, an improved version of LENs 
whose results clearly show that \len{} explanations outperform both LEN and LIME on different metrics (sensitivity and faithfulness). Moreover, a user study demonstrated that the logic explanations are more useful than the importance vector and provide a better user-experience (particularly on global explanation).
This has wide-ranging impact, as LIME is a popular strategy used in various fields (e.g., \citet{lime-use-finance} and \citet{lime-use-medicine}).

\section{Limitations}
Regarding the aggregation of local explanations, the proposed algorithm can be intractable in case $k$ is not small. To alleviate this issue, we are working on a selective algorithm to automatically filter out the local explanations that are less useful for the task.
Furthermore, since LENs require concepts as input, we did not consider models taking sequential input in this work.
In future work, we will test the explanation of the proposed model when explaining sequential models, making use of concept extraction from sequential models, like the work
done by \citet{latent-concepts-bert}. 
The backbone itself of the LEN is an MLP architecture, but it might be interesting to devise a LEN-version of an RNN or a Transformer model.
The human survey does represent the target users, as the topic experts for StackOverflow questions are computer scientists. However, in future work, to better represent the population of possible users, we aim at expanding the portion of not expert in neither Machine learning nor Computer Science. 
Finally, the paper only compares LEN and LIME explanation on one dataset, but it might be interesting to broaden the comparison to include SHAP, LORE and Anchors, while considering a variety of datasets.

\section*{Acknowledgments} 
GC acknowledges support from the EU Horizon 2020 project AI4Media, under contract no. 951911 and by the French government, through Investments in the Future projects managed by the National Research Agency (ANR), 3IA Cote d’Azur with the reference number ANR-19-P3IA-0002.
FG is supported by TAILOR, a project funded by EU Horizon 2020 research and innovation programme under GA No 952215. This work was also partially supported by HumanE-AI-Net a project funded by EU Horizon 2020 research and innovation programme under GA 952026.

\bibliography{custom}
\bibliographystyle{acl_natbib}

\clearpage

\appendix

\section{Algorithms}
\label{app:alg}
In this section, we report the local and global explanation methods from \len{}. In particular, Algorithm \ref{alg:local-exp-imp} reports the pseudocode for improving the local explanations from the LEN, while Algorithm \ref{alg:global-exp-imp} reports the optimal aggregation mechanism proposed in this paper. 

\begin{algorithm}[h]
    \caption{\len{} Local Explanation}
    \label{alg:local-exp-imp}
    \DontPrintSemicolon
    \SetKwFunction{FLocImp}{local\_explanation}
    \SetKwProg{Fn}{Function}{:}{}
    \Fn{\FLocImp{$model$, $x$, $target\_class$}}{
    $exp \gets$ Original LEN local explanation \; 
    $org\_pred \gets$ $model(x)$\;
    $good\_terms, bad\_terms \gets [\,], [\,]$\;
    \ForAll{$term \in$ exp}{
        $x^\prime \gets$ Clone $x$ with $term$ flipped\;
        $pert\_pred \gets model(x^\prime)$\;
        \eIf{$org\_pred \leq pert\_pred$}{
            $bad\_terms$.append($term$)\;
        }{
            $good\_terms$.append($term$)\;    
        }
    }
    \KwRet Conjunction($good\_terms$)\;
    }
\end{algorithm}

\begin{algorithm}[h]
    \caption{\len{} global explanation}
    \label{alg:global-exp-imp}
    \DontPrintSemicolon
    \SetKwFunction{FAgg}{aggregate\_explanations}
    \SetKwProg{Fn}{Function}{:}{}
    \Fn{\FAgg{ $topk\_explanations$}}{
        $all\_comb \gets$ PowerSet($topk\_explanations$)\;
        \ForAll{$exps \in all\_comb$}{
            $cur\_exp \gets \bigvee exp \text{ for } exp \in exps$\;
            $accuracy \gets$ Calculate accuracy of $cur\_exp$ on $validation\_data$\;
            \If{$accuracy > best\_acc$}{
                $best\_exp \gets cur\_exp$\;
                $best\_acc \gets accuracy$\;
            }
        }
        \KwRet $best\_exp$\;
    }
\end{algorithm}

\section{Model Evaluation}
\label{sec:model-eval}
In this section, we employ the LEN directly as a classifier, to assess the performance drop required to employ an explainable-by-design network instead of a black-box one.

\begin{figure}[t]
      \centering
      \includegraphics[width=1\linewidth]{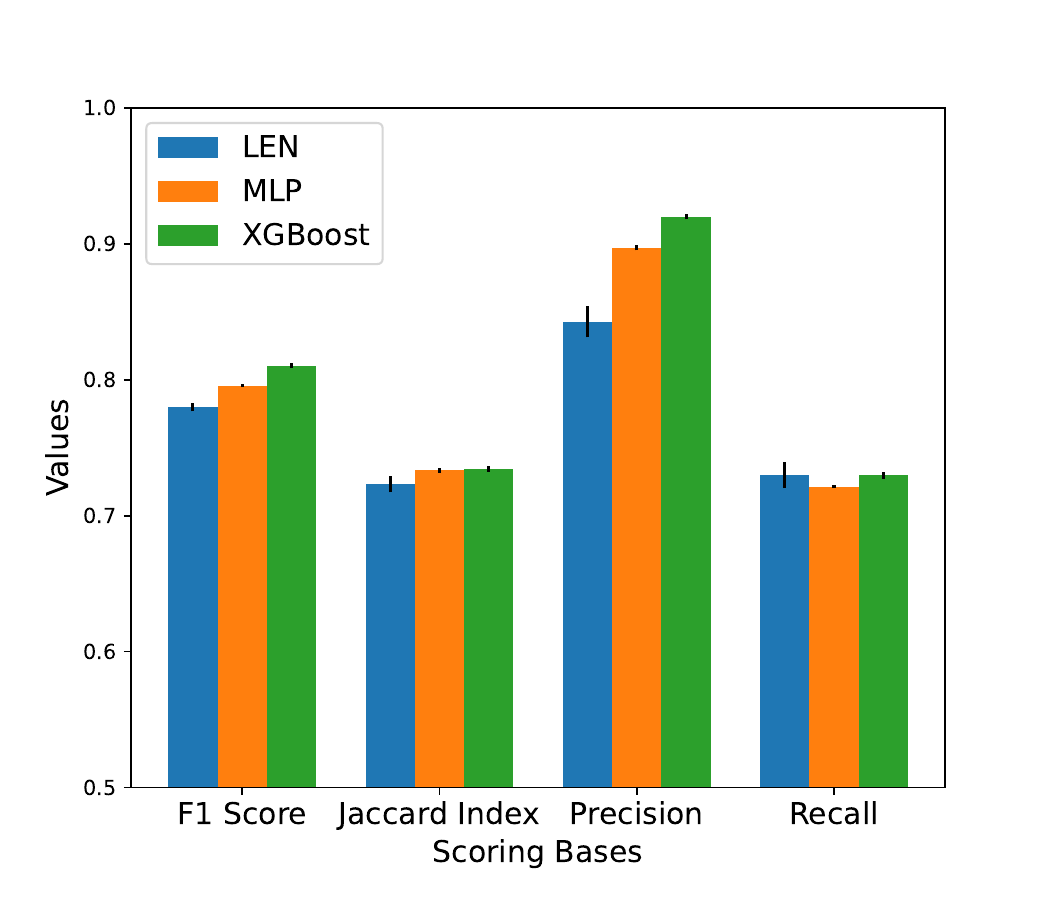}
      \caption{Evaluation Values for the different models}
      \label{plot:eval-models}
\end{figure}
Figure \ref{plot:eval-models} compares the predicting performance of the LEN with an MLP and an XGBoost, two black-box models, in terms of F1 Score, Jaccard Index, Precision and Recall. 
Results are averaged 10 times over different test-train splits and model initializations. We also report the $95 \%$ confidence intervals.

We observe that XGBoost performs better than both the LEN model and the MLP in all metrics. 
We can also appreciate that the LEN proposed in \citep{barbiero2022entropy} only slightly decreases the performance w.r.t. using almost identical MLP. A higher difference was expected, as in general there exists a trade-off between the model explainability and its performance \citep{xai-performance-tradeoff}.
These results indicate that 
the performance of LEN is good/comparable enough to consider replacing outperforming black-box models to gain higher interpretability. 

\section{Experimental details}
\label{sec:details}
\paragraph{Hardware}
All experiments were run on a machine equipped with an Intel i7-8750H CPU, an NVIDIA GTX 1070 GPU and 16 GB of RAM. 

\paragraph{Hyper-parameters} The selection was done with a grid search alongside, to maintain fairness in comparison, a constraint on the time required to obtain explanations.

\paragraph{Simulation Experiments} The details about the different settings, S1 and S2, of the experiment described in Section \ref{sec:sim-exps}, is as follows: In each run
of S1, we add 2 noisy features. In training data, each noisy features is added with a 30\% probability of being added to inputs of tag C\# and 5\% probability to the other tags.
In testing data, it is added uniformly added to all tags with 5\% probability. Bias is ensured by having a threshold difference of 0.03 between the test and validation F1 scores.

S2 follows similarly, where we add 2 noisy features, but increase the probability of adding them to inputs of tag C\# in training data from 30\% to 35\%. The threshold difference
of F1 scores is also increased to 0.05. This is done to get a model that uses the noisy features with higher importance than that we get with setting S1.

\section{Evaluation of Trust in Explanations}
In general, trust in the explanations refers to reliability of the explanations. In this paper, we used two metrics to measure this trust in the explanations, the AUC-MoRF and the Max-Sensitivity, for which we reported the details below.

\paragraph{Area Under the Most Relevant First Perturbation Curve}
\label{sec:aucmorf}
Area Under the Most Relevant First Perturbation Curve (AUC-MoRF) \citep{auc-morf-2} is a metric based on the MoRF perturbation curve as proposed by Samek et al. \citep{auc-morf}.
MoRF curve is the plot of prediction from model versus the number of features perturbed, where the features are perturbed in a most relevant first order.
Thus, AUC-MoRF can be defined as:
\begin{equation}
    \text{AUC}_\text{MoRF}(\Phi, f, x) = \sum_{k = 2}^{D} \frac{f(y^{(k-1)}) + f(y^{(k)})}{2} 
\end{equation}
Here $f$ is the model being explained, $x$ is an input vector, $\Phi$ is an explanation method, $D$ is the number of input features and $y^{(k)}$ is the input vector
after the $k^\text{th}$ MoRF perturbation. MoRF perturbations are defined recursively as below:
\begin{equation}
\begin{split}
        y^{(0)} &= x \\
    \forall 1 \leq k \leq D : y^{(k)} &= g(y^{(k-1)}, r_k
\end{split}
\end{equation}
Here $g$ is a function that takes a vector and an index, and perturbs the given vector at the given index, and $[r_1, r_2, \ldots, r_D]$ are the indices of the input
features sorted in descending order of their relevance, as determined by the explanation $\Phi$.

In our evaluation, we normalize the AUC-MoRF values to be in the $[0, 1]$ range, by dividing the values by $D - 1$ when $D > 1$. So, the final formula used looks
like:
\begin{equation}
\begin{split}
    \text{Normalized}& \text{ AUC}_\text{MoRF}(\Phi, f, x)=    \\=  & \frac{1}{(D - 1)} \sum_{k = 2}^{D} \frac{f(y^{(k-1)}) + f(y^{(k)})}{2} 
\end{split}    
\end{equation}

A lesser value of AUC-MoRF means a more faithful explanation, and thus a more trustworthy explanation.

\paragraph{Max-Sensitivity}
\label{sec:max-sens}
Sensitivity of an explanation measures the proneness of the explanation to be affected by insignificant perturbations to the input.
Max-Sensitivity is a metric due to Yeh et al. \citep{max-sensitivity} which is defined as below:
\begin{equation}
\begin{split}
\text{SENS}_\text{MAX}&(\Phi, f, x, r) = \\
& = \max_{\left\lVert y - x \right\rVert \leq r} \left\lVert \Phi(f, y) - \Phi(f, x) \right\rVert 
\end{split}
\end{equation}
Here $f$ is the model being explained, $x$ is an input vector, $y$ is the input vector with some perturbations, $r$ is the max perturbation radius, and
$\Phi$ is an explanation method, which takes a model and input vector and gives the explanation.

The lesser the value of this metric, the lesser is the explanation prone to minor perturbations in the input, and so more is our trust in the explanation. 

\section{Human Survey}
\label{sec:appendix-survey}
In the following pages, we report a compressed copy of the human survey for which we reported the results in Section \ref{sec:eval-survey}.

In the survey section where we aim to establish the usefulness of the explainers in getting a more general classifier, we train a radial basis function (RBF) SVM to
mistakenly learn to associate the feature ‘add’ with the C\# tag. This was done by randomly adding (with 50\% probability) “add” only to the training data labeled with the C\# tag.
RBF-SVM was trained on this perturbed data, getting a 6\% smaller Jaccard Index validation score than the training one. This difference confirmed that the model mistakenly
learned to associate “add” with C\# tag.

\clearpage


\section*{Supplementary material}

# Survey comparing explanations

We have a neural network trained to predict tags for StackOverflow questions. We want to understand what rules or features this neural net model is learning and using to make its predictions. This survey aims to compare two different techniques of explaining the reason about why the predictions were made by the model. We shall name these techniques "Explainer A" and "Explainer B".

Both these techniques take the model, input and the resulting prediction. They use these to output the explanation as to why we get the prediction from the model when we pass it the input.

The survey consists of 3 sections (including this one) with questions asking you to rate, compare and infer from different explanations. This section is mostly to give a brief introduction on the explanation methods.

*Required

1. How much experience do you have with Machine Learning/Computer Science?

   *Mark only one oval.*

   ◯ I work on or study Machine Learning

   ◯ I work in or study Computer Science

   ◯ What is Machine Learning/Computer Science?

Before we start comparing the explanations, below we show how to interpret the explanations from the explainers.

Here's a sample StackOverflow question. You can notice that it has "python" as a tag. Putting this question through our trained Neural Net model, we get the predicted tag to be 'python'.

**Python importing**

Asked 11 years, 11 months ago    Active 11 years ago    Viewed 15k times

3

I have a file, `myfile.py`, which imports `Class1` from `file.py` and `file.py` contains imports to different classes in `file2.py`, `file3.py`, `file4.py`.

In my `myfile.py`, can I access these classes or do I need to again import file2.py, file3.py, etc.?

Does Python automatically add all the imports included in the file I imported, and can I use them automatically?

python    import

Explainer A (with more information below the plot)

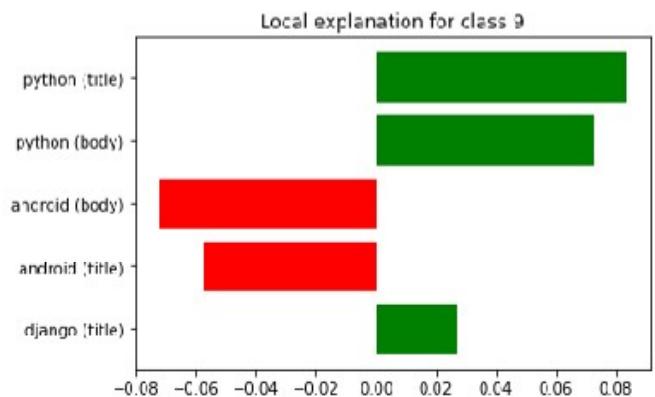

The image above shows the explanation by Explainer A. It is telling us why the model predicted the tag to be 'python'. The explanation is in the form of a graph, with longer bars meaning more important. We see that 'python (title)' and 'python (body)', i.e. python in the title and body of the question respectively, push the prediction towards 'python' as the tag. The presence of 'android' in the body or title would've pushed the prediction away from 'python' tag, but they are not present in the above example and so their absence acts in pushing the prediction towards 'python'.

2. How easy is it for you to understand why the model predicted the tag to be "Python" using the explanation by Explainer A below? (1 being the worst and 5 being the best) (Note the question does NOT ask you to rate the correctness of the model/explanation) *

   *Mark only one oval.*

   ◯ 1 – Very Hard – Very hard to infer anything from the explanation; doesn't help at all
   ◯ 2 – Hard
   ◯ 3 – Okay
   ◯ 4 – Easy
   ◯ 5 – Very easy – Extremely helpful and easy-to-understand explanations, and I understand why this prediction was made

Explainer B (with more information below the formula)

$$\neg android(body) \land python(body) \land \neg android(title) \land \neg c\#(body) \land \neg php(body)$$

The formula above shows the explanation by Explainer B. It is also telling us why the model predicted the tag to be 'python'. The explanation tells us that we got this prediction because we have 'python' in the body of the question, we don't have 'android' in the title or the body and we don't have 'c#' or 'php' in the body of the question. Note that "∧" refers to conjunction/"and" (&).

3. How easy is it for you to understand why the model predicted the tag to be "Python" using the explanation by Explainer B below? (1 being the worst and 5 being the best) (Note the question does NOT ask you to rate the correctness of the model/explanation) *

   *Mark only one oval.*

   ◯ 1 – Very Hard – Very hard to infer anything from the explanation; doesn't help at all
   ◯ 2 – Hard
   ◯ 3 – Okay
   ◯ 4 – Easy
   ◯ 5 – Very easy – Extremely helpful and easy-to-understand explanations, and I understand why this prediction was made

Note that both the explainers have told us similar things but in different ways.

Improving Model

We are now only interested in 3 tags – C#, Java and Python. Neural Net models often learn wrong rules/patterns. We want you to look at the general predictions by the two explainers to find a rule/pattern that you think is not logically correct.

For example, suppose the explanations suggests that the model is predicting the tag "Python" because the question has the word "file" in it. However, we know that "file" is equally likely to be used in C# and Java. So, this would be one such rule/feature that is wrongly being learned by the model.

Explainer A

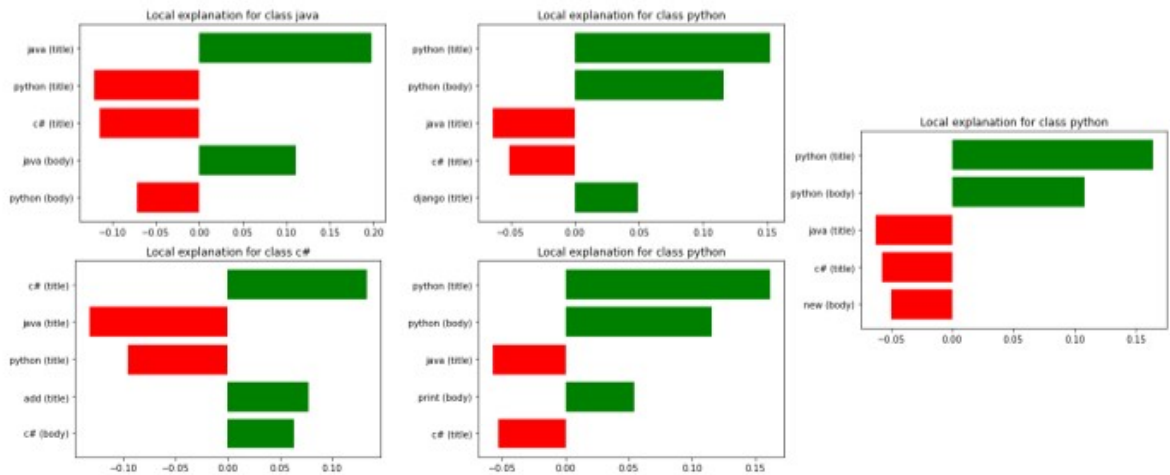

The Explainer A runs multiple inputs through the model and presents a carefully selected subset of them to give explanations for, so that we can understand what the model is doing in general. That is why you see 5 plots in the above image, each showing explanation for different inputs.

4. Using the explanation from Explainer A, which input term/feature (like "sort (title)") do you think we should ignore to get a more generalized model? (put "None" if you can't find any such feature) *

_____

5. How easy was it to understand the explanation by Explainer A and choose the term to ignore? *

Mark only one oval.

|  | 1 | 2 | 3 | 4 | 5 |  |
|---|---|---|---|---|---|---|
| Very Hard | ○ | ○ | ○ | ○ | ○ | Very Easy |

## Explainer B

### Explanations by Explainer B

$$add(title) \lor c\#(title) \implies c\# \text{ tag}$$
$$eclipse(title) \lor java(title) \lor java(body) \implies java \text{ tag}$$
$$django(title) \lor python(title) \lor python(body) \implies python \text{ tag}$$

Explainer B gives a logic formula with general rules that are followed to get the tags. There are 3 formulae above, one for each tag. Note that "V" refers to disjunction/"or" (|).

6. Using the explanation from Explainer B, which input term/feature (like "sort (title)") do you think we should ignore to get a more generalized model? (put "None" if you can't find any such feature) *

_____

7. How easy was it to understand the explanation by Explainer B and choose the term to ignore? *

   Mark only one oval.

   |   | 1 | 2 | 3 | 4 | 5 |   |
   |---|---|---|---|---|---|---|
   | Very Hard | ○ | ○ | ○ | ○ | ○ | Very Easy |

## Comparing Classifiers

We have two models. We want to choose one of them. So, we use our explainers to give some insight on the models.

Below you will see general explanations for both the models. Look at the explanations below and see which model seems more correct.

## Explainer A

### Explanations of Model 1 by Explainer A

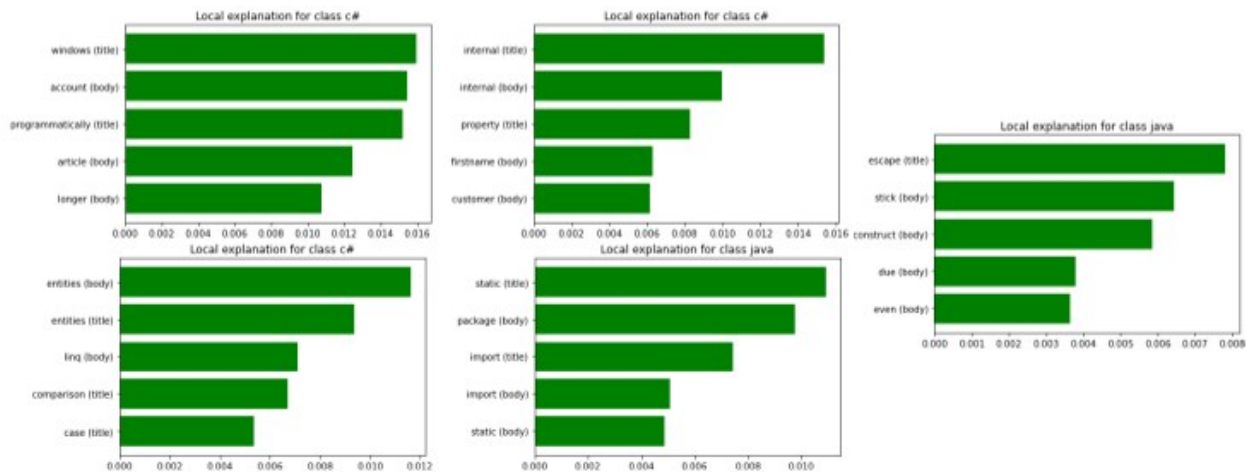

Explanations of Model 2 by Explainer A

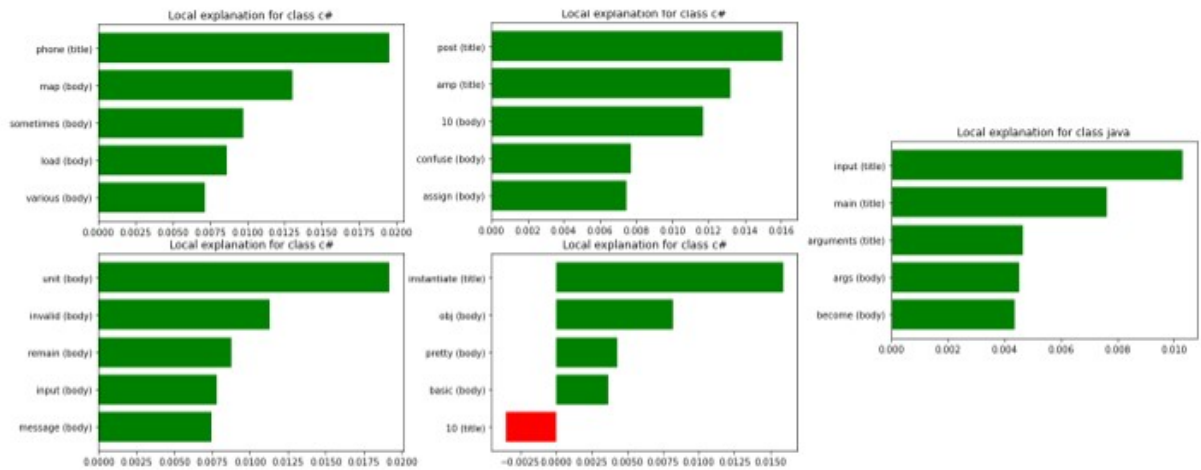

8. With reference to the explanations by Explainer A, which model do you think would perform better in the real world? *

   Mark only one oval.

   ◯ Model 1
   ◯ Model 2
   ◯ Can't decide. Both seem equally good/bad

9. Optionally, could you please provide the reasoning behind your previous answer?

   _____

10. How easy was it for you to compare the models using the explanations from Explainer A? (1 being the worst and 5 being the best) *

    Mark only one oval.

    ◯ 1 – Very Hard – Very hard to infer anything from the explanation; doesn't help at all
    ◯ 2 – Hard
    ◯ 3 – Okay
    ◯ 4 – Easy
    ◯ 5 – Very Easy – Extremely helpful and easy-to-understand explanations

Explainer B

Explanations of Model 1 by Explainer B

$$c\#(title) \implies c\# \text{ tag}$$
$$android(title) \lor java(title) \lor jar(body) \implies java \text{ tag}$$
$$python(title) \lor python(body) \implies python \text{ tag}$$

Explanations of Model 2 by Explainer B

$$c\#(title) \lor var(body) \implies c\# \text{ tag}$$
$$best(title) \lor java(title) \lor java(body) \implies java \text{ tag}$$
$$python(title) \lor python(body) \implies python \text{ tag}$$

11. With reference to the explanations by Explainer B, which model do you think would perform better in the real world? *

    *Mark only one oval.*

    ◯ Model 1

    ◯ Model 2

    ◯ Can't decide. Both seem equally good/bad

12. Optionally, could you please provide the reasoning behind your previous answer?

    _____

13. How easy was it for you to compare the models using the explanations from Explainer B? (1 being the worst and 5 being the best) *

    *Mark only one oval.*

    ◯ 1 – Very Hard – Very hard to infer anything from the explanation; doesn't help at all

    ◯ 2 – Hard

    ◯ 3 – Okay

    ◯ 4 – Easy

    ◯ 5 – Very easy – Extremely helpful and easy-to-understand explanations

Thank you

14. Thank you for your time. If you have any comments, please mention them below.

    _____



\end{document}